\documentclass[runningheads]{llncs}
\usepackage[T1]{fontenc}
\usepackage{graphicx}
\usepackage{booktabs}
\usepackage[misc]{ifsym}

\newcommand{\corr}{(\Letter)}
\usepackage{mwe}
\usepackage{color}
\usepackage{amsmath}
\usepackage{multirow}
\usepackage{amssymb}
\usepackage{cite}

\usepackage{subfigure}

\begin{document}

\title{Propagation Structure-Semantic Transfer Learning for Robust Fake News Detection}
\toctitle{Propagation Structure-Semantic Transfer Learning for Robust Fake News Detection}

\titlerunning{PSS-TL for Robust Fake News Detection}

\author{Mengyang Chen\inst{1,2} \and
Lingwei Wei\inst{1} \corr \and
Han Cao\inst{1,2} \and Wei Zhou \inst{1} \and Zhou Yan \inst{3} \and Songlin Hu \inst{1,2}}

\tocauthor{Mengyang Chen, Lingwei Wei, Han Cao, Wei Zhou, Zhou Yan, Songlin Hu}

\institute{Institute of Information Engineering, Chinese Academy of Sciences, Haidian District Beijing 100085, China \email{\{chenmengyang, weilingwei, caohan, zhouwei, husonglin\}@iie.ac.cn}
\and
School of Cyber Security, University of Chinese Academy of Sciences, Huairou District Beijing 101408, China \email{\{chenmengyang22,  caohan22\}@mails.ucas.ac.cn}
\and
State Key Laboratory of Communication Content Cognition, People's Daily Online, Chaoyang District Beijing 100020, China \email{yanzhou@people.cn}
}

\maketitle              

\begin{abstract}
Fake news generally refers to false information that is spread deliberately to deceive people, which has detrimental social effects. Existing fake news detection methods primarily learn the semantic features from news content or integrate structural features from propagation. However, in practical scenarios, due to the semantic ambiguity of informal language and unreliable user interactive behaviors on social media, there are inherent semantic and structural noises in news content and propagation. Although some recent works consider the effect of irrelevant user interactions in a hybrid-modeling way, they still suffer from the mutual interference between structural noise and semantic noise, leading to limited performance for robust detection. To alleviate this issue, this paper proposes a novel Propagation Structure-Semantic Transfer Learning framework (PSS-TL) for robust fake news detection under a teacher-student architecture. Specifically, we design dual teacher models to learn semantics knowledge and structure knowledge from noisy news content and propagation structure independently. Besides, we design a Multi-channel Knowledge Distillation (MKD) loss to enable the student model to acquire specialized knowledge from the teacher models, thereby avoiding mutual interference. Extensive experiments on two real-world datasets validate the effectiveness and robustness of our method.

\end{abstract}

\section{Introduction}

Fake news generally refers to false information that is spread deliberately to deceive people   \cite{shu2017fake}. 
{The rapid advancement of online media} has sparked a rise in fake news, which inflicts significant harm on society   \cite{fisher2016pizzagate,vosoughi2018spread,faris2017partisanship}, making fake news detection into a research hotspot.

\begin{figure}[t]
    \centering
    \includegraphics[width=1\linewidth]{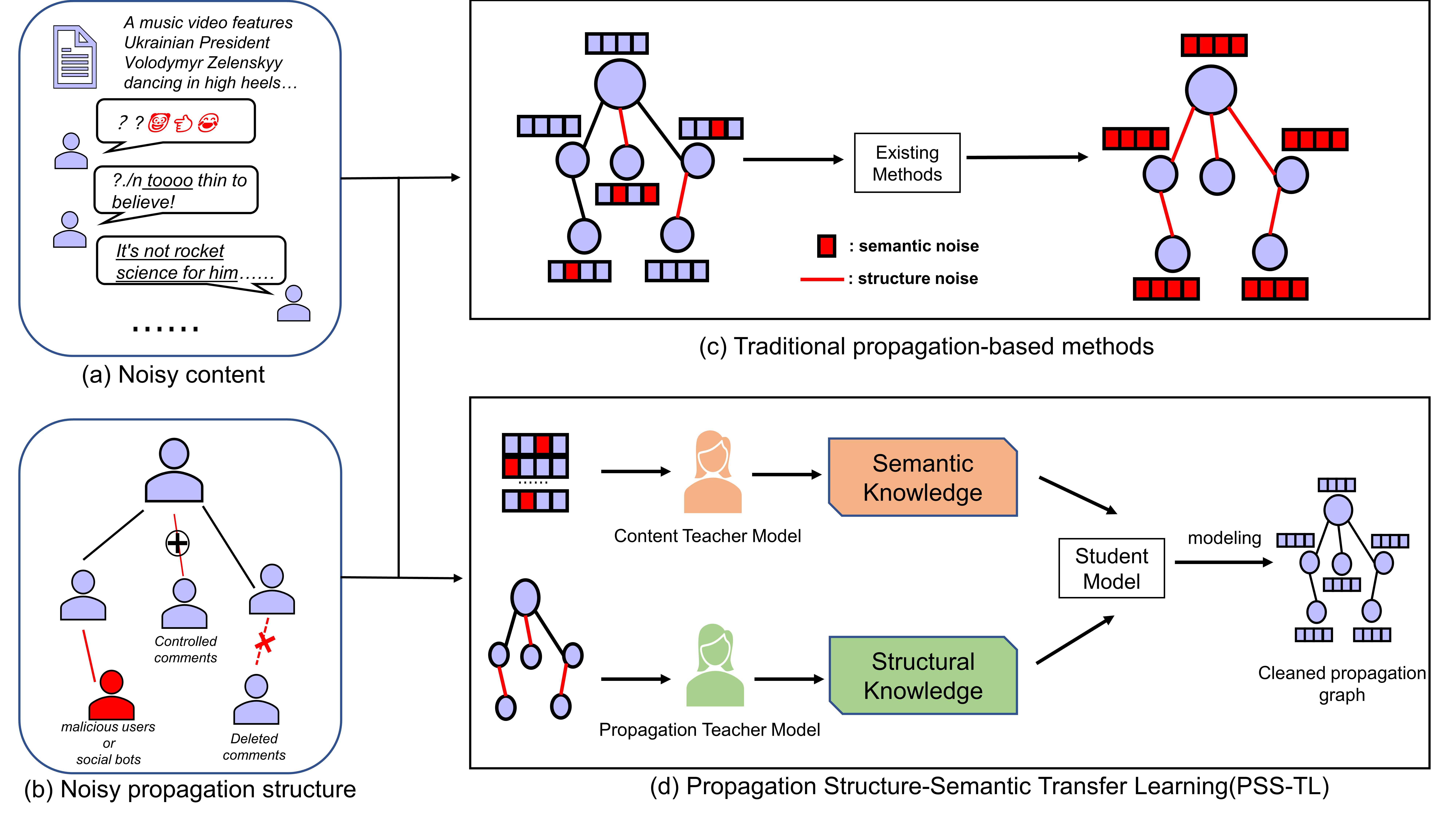}
    \caption{
    The motivation of this paper. (a): Noisy content, including garbled characters, spelling errors, and idioms, contributes to semantic noise. (b)Unreliable interactions among users lead to structural noise in news propagation trees. (c) Previous works generally learn high-level features in a hybrid way. They would suffer from the mutual inference between the learning of noisy contents and incomplete propagation trees, leading to unreliable node representations. (d) Our PSS-TL adopts a dual teacher-student distillation framework, where two teachers independently transfer reliable semantic and structural features to the student model for robust detection.}
    \label{fig:noise}
\end{figure}

Existing methods of fake news detection mainly focus on textual content such as news text and contexts   \cite{castillo2011information,ma2015detect,yu2017convolutional} and propagation information such as interactions between users   \cite{ma2018rumor, Liu2018EarlyDO,bian2020rumor,song2021temporally, dou2021rumor,wei-etal-2021-towards}.  Recent works make attempts to achieve robust detection by considering the effect of noise in practical scenes. There are inherent noises in news and its corresponding propagation trees. Due to the semantic ambiguity inherent in informal language, spelling errors, and garbled characters, there is semantic noise in the news and its comments (see Fig.~\ref{fig:noise}(a)). Moreover, unreliable interactive behaviors among potential malicious users and social bots, deleted or controlled comments and limited collection of propagation data may also bring structural noise in propagation trees    \cite{ma2019detect,yang2021rumor} (see Fig.~\ref{fig:noise}(b)) \footnote{
Here, we divide the propagation graph of news into two parts: the content being spread (the content of news and comments in the propagation graph) and the propagation structure (the interactive relationships between news and comments).}. These studies mainly focus on the impact of noises in news propagation graphs and have made efforts to solve this problem through refining propagation structures   \cite{wei-etal-2021-towards, wei2022uncertainty, wu2023decor} or training a robust detection model through contrastive learning or adversarial learning   \cite{he2021rumor,sun2022rumor,ma2022towards}.

However, in extracting high-level semantic and structural features through hybrid modeling, these models confront a significant challenge: the \textbf{mutual interference} arising between semantic and structural noise. Semantic noise introduces inaccuracies in feature representation, thereby skewing the computation of structural weights. In turn, structural noise, perpetuated by flawed adjacency relationships, infiltrates the feature-learning process, compounding the disruptive effects.

Therefore, the representations learned by these models are prone to containing unreliable knowledge. As shown in Fig.~\ref{fig:noise}(c), during the modeling of propagation graphs, the noise in both nodes and edges, representing textual content and user behaviors respectively, will mutually impact each other and cause an error accumulation after the message passing. It will further make the model learn unreliable representations, i.e., most node embeddings may be poisoned, which will result in the limited performance of detection.

In this paper, we propose a novel Propagation Structure-Semantic Transfer Learning framework (PSS-TL) to alleviate the mutual interference between structural noise and semantic noise for robust fake news detection under the teacher-student architecture. 
 Different from the above methods, we separately learn semantics and propagation through dual-channel transfer learning (see Fig.~\ref{fig:noise}(d)). 
 The separate modeling by two teachers would successfully prevent interaction between noisy semantic features and structural features, therefore alleviating the mutual interference between the extraction of semantic and structural features. The student model could receive reliable knowledge provided by authoritative teachers for better fake news detection.

Specifically, we design dual-teacher models to separately learn semantic and structure knowledge from the news content and the corresponding propagation trees. 
The content teacher captures effective semantic knowledge from news content using a multi-layer perceptron. Meanwhile, the propagation teacher extracts propagation knowledge by traversing a global propagation graph of the news, which is based on common users between the news and the positional encoding vectors of the news, utilizing graph convolutional networks.
For the student model,  we design a local-global propagation interaction module to fully utilize both local and global propagation information of news. It first integrates features by modeling local interactions in each propagation graph. Subsequently, it spreads this representation across the global graph to obtain a combined representation that integrates local and global propagation information.

To better transfer reliable semantic knowledge and propagation structure knowledge to the student model for detection, we design the Multi-channel Knowledge distillation (MKD) loss, which consists of two components: teacher supervision loss and targeted guidance loss. 
The teacher supervision loss consists of distillation losses between the logits generated by the teacher model and the student model, which allows teacher models to provide specialized knowledge to the student model. Targeted guidance loss includes alignment losses between the latent representations of two teacher models and the student model. It aims to maximize the consistency of representations of the same news between each teacher model and student model, making knowledge transfer from teacher models to student models more efficient.
In this way, the student model can learn structure and semantic knowledge separately, thereby alleviating the mutual interference between structural noise and semantic noise.

We conduct experiments on two real-world public fake news datasets. The experimental results
show that our PSS-TL significantly outperforms comparison baselines and achieves better detection performance, indicating the effectiveness and superiority of our method. 
Extensive experiments further indicate better generalization ability across domains and promising robust detection performance under various noisy scenarios.

The contributions of this work can be summarized as follows:
1) We propose a novel Propagation Structure-Semantic Transfer Learning framework (PSS-TL) for improving robust fake news detection. It alleviates the mutual interference between structural noise and semantic noise in propagation modeling via a dual teacher transfer learning network.
2) We design a new multi-channel knowledge distillation loss for better training the student model to pertinently transfer reliable structure and semantic knowledge to the student network.
3)  Experimental results demonstrate the effectiveness and superiority of PSS-TL for fake news detection. Our PSS-TL obtains better robust detection performance under various noise scenarios \footnote{The code is available at github.com/IMCMY99/PSS-TL.}.

\section{Related Work}

The goal of detecting fake news is to identify and assess the authenticity of a piece of information. Existing methods for detecting fake news mainly focus on two aspects: textual content and news propagation.

\textbf{Text-based fake news detection methods} 
extract semantic patterns from news textual content for detection through the employment of feature engineering   \cite{castillo2011information,popat2017assessing,ma2015detect} and a wide array of deep learning architectures, including neural networks   \cite{ruchansky2017csi, karimi2019learning} and pre-trained language models    \cite{kaliyar2021fakebert,jwa2019exbake}.
 Some works also integrate tasks such as stance detection and sentiment analysis with fake news detection, enabling multi-task learning  \cite{luvembe2023dual,hamed2023fake}.

\textbf{Propagation-based fake news detection methods} capture the propagation patterns of news by modeling the interactions between news and comments into time series   \cite{ma2016detecting,Liu2018EarlyDO} or topological structures such as propagation trees   \cite{ma2018rumor,dou2021rumor} and propagation graphs   \cite{bian2020rumor,wei-etal-2021-towards,wei2022uncertainty}.  Some studies further explore multi-relational interactions between the users and news in the propagation graph   \cite{yuan2020early,dou2021user}.

Recently, to alleviate the incomplete propagation issue, some works enhance the robustness of Graph Neural Network (GNN) models and augment the original propagation graph. Studies such as   \cite{he2021rumor,sun2022rumor,ma2022towards} propose different augmentations to the propagation structure for graph contrastive learning to enhance the robustness of GNN models. Others, like   \cite{wei-etal-2021-towards,wei2022uncertainty,wu2023decor,9837882}, optimize the original propagation graph through specific objective functions. 
However, these methods fail to consider the mutual interference between structural noise and semantic noise when modeling real-world propagation, as semantic noise spreads through noisy propagation structure and further intensifies the issue of noise in propagation, resulting in the limited performance of detection.

\section{Propagation Structure-Semantic Transfer Learning Framework}

\begin{figure*}[t]
    \centering
    \includegraphics[width=1\linewidth]{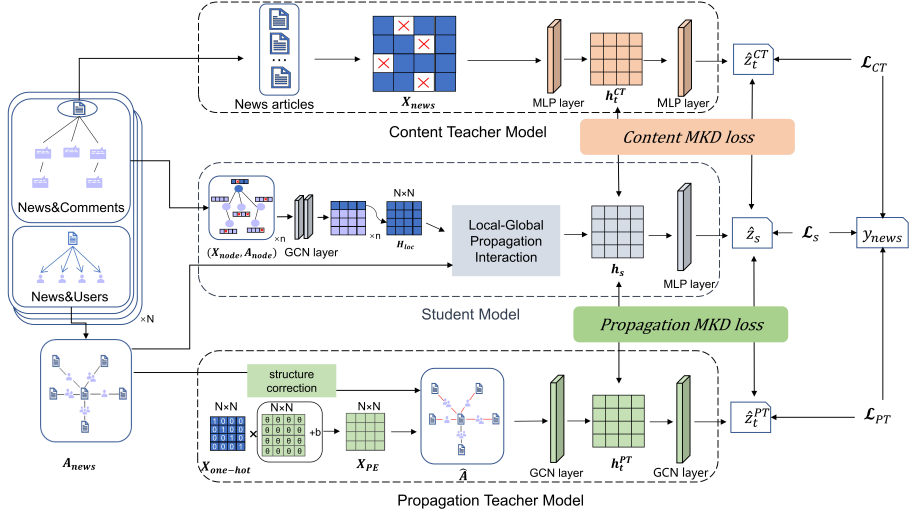}
    \caption{ The overall architecture of PSS-TL.}
    \label{fig:PSS}
\end{figure*}

In this section, we propose a novel Propagation Structure-Semantic Transfer Learning framework (PSS-TL), to independently address noise in both semantics and propagation structures, thereby avoiding mutual interference between them.

\paragraph{Problem Statement}
Fake news detection is to verify the authenticity of a given news article, we take fake news detection as a binary classification problem,  where each sample consists of news, comments, users, propagation structures, and user engagements, with each sample annotated with a ground truth label indicating its authenticity. Formally, Dataset $\mathcal{D}$ consists of $N$ samples and each sample is represented by $\mathcal{G} = (\mathcal{V,E,U_{\mathcal{G}},S_{\mathcal{G}}})$, where $\mathcal{V} = \{r,c_1,...,c_N\}$ represents the features of the news $r$ and its comments ($c_1,...,c_N$),  $\mathcal{U_{\mathcal{G}}}$ represents the users engaged in the news, $\mathcal{E}$ represents a set of explicit interactive behaviors, e.g., retweet, and $\mathcal{S_{\mathcal{G}}}=\{s_u, u\in \mathcal{U_{\mathcal{G}}}\}$ represents the engagements of users in news, in which $s_u\in \mathcal{S_{\mathcal{G}}}$ is the number of times user $u$ interact with news.
The task objective of fake news detection is to learn a classifier $f$ to classify samples and determine whether the news is true (labeled as 0) or false (labeled as 1), i.e., $$f:\mathcal{G} \longrightarrow y, \quad y \in \{0,1\}.$$

\subsection{Overview}
As shown in Fig.~\ref{fig:PSS}, PSS-TL has three components: a content teacher model, a propagation teacher model with structure correction and a local-global propagation interaction module. The content teacher captures semantic knowledge through a multi-layer perceptron.
The propagation teacher extracts structural knowledge from a global perspective through the common participating users among news, with the structure correction mechanism which is based on the frequency of common users.
The Local-Global Propagation Interaction (LGPI) module is designed to fully integrate local propagation information and global propagation information. 
We design the Muliti-channel Knowledge distillation (MKD) loss for model training.
Through MKD loss, the professional knowledge in the teacher models is fully independently extracted to the student model, allowing the student model to learn semantic and structural knowledge separately, avoiding mutual interference between semantics and propagation structures.

\subsection{Dual Teacher Models}
As there will be mutual interference of semantic noise and structural noise when modeling content and propagation jointly, we adopt dual-teacher models to learn semantic and structural knowledge from content and propagation containing noise, respectively.
\subsubsection{Content Teacher Model}
We utilize multi-layer perceptron as the backbone of the content teacher model for automatic semantic feature learning from noisy content. Given the input content $X_{news}$, the hidden representation $h_{t}^{CT}$ and output $\hat{z}_{t}^{CT}$ can be computed as,
\begin{equation}
\begin{aligned}
h_{t}^{CT} &= \text{MLP}_{1}(X_{\text{news}}) \\
\hat{z}_{t}^{CT} &= \text{ReLU}(\text{MLP}_{2}(h_{t}^{CT})).
\end{aligned}
\end{equation}

\subsubsection{Propagation Teacher Model}

Inspired by Yuan et al.\cite{rumor_yuan_2019}, we construct a global propagation graph $A_{\text{news}}\in \mathbb{R}^{N\times N}$ based on the shared participation of users among news items and refine it to learn structural knowledge, which can effectively reduce computational resource consumption.

Firstly, we construct a user-engagements graph $E\in \mathbb{R}^{N\times |\mathcal{U}|}$, where $\mathcal{U}=\cup_{\mathcal{G}}\mathcal{U}_\mathcal{G}$ contains all users in the dataset $\mathcal{D}$ and $E_{i,j} $ reprensents the number of times users $u_j$ participate in news $\mathcal{G}_i$. And then we get the global propagation graph $A_{\text{news}}$ by:

\begin{equation}
A_{\text{news}}=EE^{\prime}.
\end{equation}

Then we design a positional encoding vector to avoid the interference from semantic noise on structure learning. 
Formally, for each sample, we define a learnable propagation vector using a one-hot positional encoding vector $X_{\text{one-hot}} \in \mathbb{R}^{n \times n}$ as follows:
\begin{equation}
    X_{PE}=W X_\text{one-hot}+b ,
\end{equation}
where $W$ is a learnable weight matrix and $b$ is a learnable bias vector.

Some studies have found that the common users between news are related to the veracity of the news pairs. inspired by  Karrer et al. \cite{karrer2011stochastic} and Newman \cite{newman2018network}, we employ a multi-layer perceptron to get the refined global propagation structure $\hat{A}_{\text{news}}$ by increasing the weight of the edges between news items of the same veracity and decreasing those of different veracity in the original global propagation structure $A_{\text{news}}$.
The formal computation is as follows:
\begin{equation}
   \hat{A}_{\text{news}}=A_{\text{news}}·M+I ,
\end{equation}
where $I$ is the identity matrix and $M$ is the edge retention matrix, calculated based on the degrees of the pairwise propagating nodes:

\begin{equation}
\begin{aligned}
M_{ij} &= \begin{cases} 
             \text{Softmax}\big(\text{MLP}(A_{\text{news},ij},d_i,d_j)\big) & \text{if } A_{\text{news},ij} \neq 0 \\
             0 & \text{else} 
         \end{cases}.
\end{aligned}
\end{equation}

Given the propagation vector $X_{PE}$ and $\hat{A}_{\text{news}}$, we utilize a two-layer graph convolutional network (GCN) to compute the hidden representation $h_{str}$ of all news and the output $\hat{z}_{str}$:

\begin{equation}
\begin{aligned}
    h_{t}^{PT}&=\text{GCN}_{1}(\hat{A}_{\text{news}},X_{PE}) \\ 
    \hat{z}_{t}^{PT} &=\text{ReLU}(\text{GCN}_2(h_{t}^{PT})).
\end{aligned}
\end{equation}

\subsubsection{Training of Dual Teacher Models}
We train the above teacher models by using the Cross-Entropy classification loss of the predicted results outputted by the teacher models. The objective can be defined as follows:
\begin{equation}
    \mathcal{L}_{tea}=\text{CE}(\hat{z}_{t}^{tea},y_{\text{news}}),\quad tea\in\{{CT}, {PT}\},
\end{equation}
where $\text{CE}( \cdot )$ denotes the cross-entropy loss, and $y_{\text{news}} \in \{0, 1\}$ represents the ground-truth label of the news.

\subsection{Local-Global Propagation Interaction Enhanced Student Model}

To better utilize the local and global propagation information of news, we design the Local-Global Propagation Interaction module. 

As the news propagation graphs are fed into the student model and local propagation information of all news is output, the learned local propagation information is further spread on the global propagation structure constructed by the propagation teacher through the LGPI module, as shown in equation (\ref{LGPI}). 

Taking GCN as an example, the node features $X_{node}$ of each news and its propagation graph $A_{node}$ are fed into GCN to obtain local representations $H_{loc}$. Then, the local representations $H_{loc \times n}$ of all news items and the global propagation graph $A_{news}$ are fed into the LGPI module to incorporate inter-news relationships, resulting in a representation $h_{s}$ that captures the contextual information among news. This representation is then fed into a linear layer to produce the predicted output $\hat{z}_{s}$. The specific steps are as follows:

\begin{equation}
    H_{loc}=\text{ReLU}(\text{GCN}_{loc}(X_{\text{node}},A_{\text{node}}))
\end{equation}
\begin{equation}
    h_{s}=W_1(A_{\text{news}}·H_{loc\times N})+b_1
    \label{LGPI}
\end{equation}
\begin{equation}
    \hat{z}_{s}=\text{ReLU}(W_{2}·h_{s}+b_2),
\end{equation}
where $W_1$ and $W_2$ are parameter matrices, and $b_1$ and $b_2$ are bias terms.

\subsection{Multi-channel Knowledge Distillation Training Objective}
To better transfer the reliable knowledge obtained by each teacher to the student model, we develop a novel Multi-channel Knowledge Distillation (MKD) loss to perform effective and targeted supervision in the training stage of the student model.
It mainly involves teacher supervision loss and targeted guidance loss.

\textbf{Teacher supervision loss} $\mathcal{L}_\text{sup}$ measures the distillation loss that arises from the teacher model’s supervision to the student model’s output. It quantifies the degree of alignment between the two models’ predictions, i.e.,
\begin{equation}
\begin{aligned}
    \mathcal{L}_\text{sup} = D_\text{KL} \left( \text{Softmax} \left( \frac{\hat{z}_{s}}{\rho} \right), \text{Softmax} \left( \frac{\hat{z}_{t}}{\rho} \right) \right),
\end{aligned}
\end{equation}

where $D_\text{KL} ( \cdot )$ represents the Kullback-Leibler divergence between two distributions. $\hat{z}_{s}$ and $\hat{z}_{t}$ refer to the predicted output generated by the student model and a teacher model. $\rho$ is the temperature parameter to adjust the smoothness of the output generated by the teacher models.

\textbf{Targeted guidance loss} $\mathcal{L}_\text{tar}$ serves as additional supplementation of knowledge from the teacher model to the student model. It achieves consistency between the hidden representations of the same news between the teacher model and the student model while minimizing consistency between the hidden representations of different news, i.e.,

\begin{equation}
\begin{aligned}
\mathcal{L}_\text{tar} = \frac{1}{|V|} \sum_{v_i \in V} \log \frac{e^{\langle h_{s,i}, h_{t,i} \rangle}}{e^{\langle h_{s, i}, h_{t ,i} \rangle} + \sum_{j \neq i} e^{\langle h_{s, i}, h_{t, j} \rangle}}, 
\end{aligned}
\end{equation}
where $h_{s}$ and $h_{t}$ refer to the hidden representations in the modeling of the student model and teacher model, respectively.

Thus, the total objective of MKD loss can be defined as,
\begin{equation}
    \mathcal{L}_\text{MKD} = \mathcal{L}_\text{sup} + \mathcal{L}_\text{tar} .
\end{equation}

To sum up, the total objective of training the student model can be,
\begin{equation}
\begin{aligned}
    \mathcal{L}_{s} 
     = \mathcal{L}_\text{CLS} + 
     \underbrace{\lambda\mathcal{L}_\text{sup}^{PT}
     +\beta\mathcal{L}_\text{tar}^{PT}}_{\mathcal{L}^{PT}_\text{MKD}}
     + \underbrace{(1-\lambda)\mathcal{L}_\text{sup}^{CT} +(1-\beta)\mathcal{L}_\text{tar}^{CT}}_{\mathcal{L}^{CT}_\text{MKD}} ,\\
\end{aligned}
\end{equation}
where $\mathcal{L}_\text{CLS}$ refers to the cross-entropy loss based on the predicted and ground-truth labels for fake news detection. $\lambda$ and $\beta \in [0,1]$  are two hyperparameters to control the emphasis of supervision loss and targeted guidance loss, respectively.
$\mathcal{L}^{CT}_\text{MKD}$ and $\mathcal{L}^{PT}_\text{MKD}$ are the reliable knowledge distillation losses from content and propagation teacher models to the student model, respectively.

\section{Experiment}

\subsection{Experimental Setups}
\subsubsection{Datasets}
We evaluate our method on two real-world datasets, PolitiFact and GossipCop   \cite{shu2020fakenewsnet}. PolitiFact and GossipCop include news articles from fact-checking websites PolitiFact and GossipCop. To leverage user engagement, we utilize news data collected by Dou et al.  \cite{dou2021user}. 
Besides, to achieve cross-domain detection, we utilize COAID   \cite{cui2020coaid} as the target domain for generalization evaluation, which consists of 185 claims about Covid-19 healthcare and 15,996 posts.
We divided the datasets into training, validation, and testing sets in a ratio of 7:1:2.
The statistics of these datasets are shown in Table \ref{tab:datasets}.

\begin{table}[t]
    \centering
        \caption{The statistics of datasets.     }

    \begin{tabular}{l|rrr}
    \hline 
       Datasets         & \multicolumn{1}{c}{PolitiFact}  & \multicolumn{1}{c}{GossipCop} & \multicolumn{1}{c}{COAID}   \\ 
       \hline

       Number of News              &  314 &   5,464&  186 \\  
 
       Number of True News      &      157	& 2,732 &  166\\ 
     Number of False News    &  157 &2,732 &18 \\  

       Number of Posts   &40,740   &308,798 & 15,996

     \\ 
       Number of Users    &   	30,812	& 75,914 & 11,467
  
     \\ 

        \hline 
\end{tabular}
    \label{tab:datasets}
\end{table}

\subsubsection{Comparison Methods}

We compare with the following representative fake news detection methods.

\textbf{GCN}   \cite{kipf2016semi} learns the representation of each news article by performing graph convolutional operations on the news propagation graph.
\textbf{GAT}   \cite{velivckovic2017graph} employs a Graph Attention Network to encode the propagation structure of news.
\textbf{GraphSAGE}   \cite{hamilton2017inductive} models the news propagation graph using a Graph Sampling Aggregation Network, learning the representation of each article by randomly sampling and aggregating the features of neighboring nodes. 
\textbf{Bi-GCN}   \cite{bian2020rumor} constructs a bidirectional propagation graph based on the news propagation graph to enhance the effectiveness of representation learning.
\textbf{EBGCN}   \cite{wei-etal-2021-towards} is a Bayesian graph convolutional network-based rumor detection methodology that models uncertainty within rumor propagation.
\textbf{UPSR}   \cite{wei2022uncertainty} constructs an uncertainty-aware propagation structure reconstruction graph based on the original news propagation graph, enhancing the effectiveness of representation learning through Gaussian estimation.
\textbf{UPFD}   \cite{dou2021user} combines various signals such as news
text features and user preference features by jointly modeling news content and a propagation tree using a connection. 
\textbf{DECOR}   \cite{wu2023decor} constructs a news social graph based on the engagement of news users and optimizes the social graph using a stochastic blockmodel to enhance the effectiveness of graph-based representation learning for fake news detection.

\subsubsection{Implementation Details}
All experiments were conducted on a single Tesla V100, using PyTorch version 1.12.1 and Geometric version 2.3.1. We employ a pre-trained BERT (\textit{bert-base-uncased}) to encode the textual features of each news article and comment. The dimension of the hidden vectors is set to 64, and the content teacher model utilizes a two-layer MLP, while the propagation teacher model employs a two-layer GCN. The learning rate for the propagation teacher model and student model is set to 5e-4 and that for the content teacher is set to 5e-5.
We search the optimal parameters $\lambda$ and $\beta$ from \{0.1, 0.2, 0.3, 0.4, 0.5, 0.6, 0.7, 0.8, 0.9\}, and the temperature parameter $\rho$ in \{1, 2, 5, 7, 10\}. We implement comparison methods under the same environment according to the parameter setting reported in their original papers. We run five times and report the average results.

We use accuracy and macro-average F1 score as the evaluation metrics for each model. In subsequent sub-experiments, all methods employ GCN as the backbone GNN.

\begin{table}[t]
    \centering
        \caption{Results (\%) for fake news detection on PolitiFact and GossipCop. The best result is in boldface. For each method, we run five times and report the average results.}
    \label{tab:no_mask}

        \begin{tabular}{l|cc|cc}
            \hline 
            \multicolumn{1}{c|}{\multirow{2}{*}{Models}} & \multicolumn{2}{c|}{\multirow{1}{*}{PolitiFact}} & \multicolumn{2}{c}{GossipCop} \\ 
            & Accuracy& Macro-F1 & Accuracy& Macro-F1 \\ \hline 
            \multicolumn{1}{l|}{{GCN}} & 81.63{$\pm$1.82} & 81.46$\pm$1.89 & 94.88$\pm$0.11 & 94.84$\pm$0.11  \\ 
           \multicolumn{1}{l|}{{GAT}} & 82.72$\pm$0.79 & 82.61$\pm$0.81 & 95.35$\pm$0.17 & 95.27$\pm$0.17 \\ 
           \multicolumn{1}{l|}{{GraphSAGE}} & 82.81$\pm$1.54 & 82.68$\pm$1.57 & 95.21$\pm$1.48 & 95.16$\pm$1.50\\           
           \multicolumn{1}{l|}{{BiGCN}} & 84.62$\pm$0.46 & 82.98$\pm$0.46 & 96.06$\pm$0.98 & 95.94$\pm$0.99\\ 
           \multicolumn{1}{l|}{{EBGCN}} & 89.87$\pm$3.00 & 89.29$\pm$3.16 & 96.70$\pm$0.71 & 96.75$\pm$0.69 \\ 
           \multicolumn{1}{l|}{{UPSR}} & 91.46$\pm$2.43& 90.86$\pm$2.43 & 96.86$\pm$0.61 & 96.72$\pm$0.63 \\ 
                       \hline
           \multicolumn{1}{l|}{{UPFD}} \\
            \multicolumn{1}{l|}{{\quad w/ GCN}} & 82.99$\pm$2.78 & 82.69$\pm$2.92 & 95.35$\pm$0.38 & 95.30$\pm$0.38  \\ 
            \multicolumn{1}{l|}{{\quad w/ GAT}} & 82.99$\pm$0.84 & 82.90$\pm$0.86 & 96.01$\pm$0.41 & 95.97$\pm$0.41 \\ 
            \multicolumn{1}{l|}{{\quad w/ GraphSAGE}} & 85.81$\pm$3.13 & 85.60$\pm$3.05 & 95.78$\pm$0.36 & 95.75$\pm$0.36  \\  
                        \hline
           \multicolumn{1}{l|}{{DECOR}} \\
            \multicolumn{1}{l|}{{\quad w/ GCN}} & 88.71$\pm$0.00  & 88.47$\pm$0.00 & 97.22$\pm$0.12 & 97.21$\pm$0.12  \\ 
            \multicolumn{1}{l|}{{\quad w/ GAT}} & 91.29$\pm$0.82 & 91.03$\pm$0.79 & 96.70$\pm$0.06 & 96.70$\pm$0.06 \\ 
            \multicolumn{1}{l|}{{\quad w/ GraphSAGE}} & 88.39$\pm$0.64 & 88.12$\pm$0.69 & 96.60$\pm$0.07 & 96.60$\pm$0.07  \\ 
            \hline
           \multicolumn{1}{l|}{{\textbf{PSS-TL}(Ours)}} \\
            \multicolumn{1}{l|}{{\quad w/ GCN}} & \textbf{93.23}$\pm$0.18  & 92.47$\pm$0.19 & \textbf{97.71}$\pm$0.19 & 97.62$\pm$0.20 \\ 
            \multicolumn{1}{l|}{{\quad w/ GAT}} & 93.05$\pm$0.12 & \textbf{92.94}$\pm$0.17 & \textbf{97.71}$\pm$0.16 & \textbf{97.71}$\pm$0.16 \\ 
            \multicolumn{1}{l|}{{\quad w/ GraphSAGE}} & 92.83$\pm$0.14 & 92.86$\pm$0.17 & 97.23$\pm$0.027 & 97.23$\pm$0.17 \\     \hline        
        \end{tabular}
\end{table}

\subsection{Main Results}

The overall experimental results are shown in Table \ref{tab:no_mask}. From the tables, it can be observed that under different backbones of student models, PSS-TL achieves state-of-the-art detection performance, showing the superiority of our method.

From the results, we have the following observations:

1) Some methods such as DECOR, EBGCN, and UPSR that account for noisy user behaviors in the modeling of propagation, exhibit significantly better performance than those that ignore the effect of propagation noise, such as BiGCN and basic GNNs.

2) Our method outperforms DECOR, EBGCN, and UPSR on both datasets. This can be attributed to the dual-teacher knowledge distillation architecture effectively preventing interference between the semantic and structural noises. The dual teacher models can transfer reliable knowledge to the student model for better detection.

3) When the student model adopts various GNN backbones, our PSS-TL gains promising performance consistently on both datasets. It shows the insensitivity of our framework to different GNN architectures.

\subsection{Ablation Study}

We conducted ablation experiments to evaluate the key components of our method. 
\textbf{w/o Content Teacher} refers to removing the content teacher component.
\textbf{w/o Propagation Teacher} indicates removing the propagation teacher component.
\textbf{w/o $\mathcal{L}_{\text{tar}}$} and  \textbf{w/o $\mathcal{L}_{\text{sup}}$} refer to removing the targeted guidance loss and teacher supervision loss from dual teacher models to the student model in our MKD loss, respectively. 
\textbf{w/o LGPI} means removing the Local-Global Propagation Interaction module in the student model.

\begin{table}[t]
    \centering
        \caption{Results (\%) of ablation study. The best result is in boldface. }
    \label{tab:ablation}

        \begin{tabular}{l|cc|cc}
            \hline 
            \multicolumn{1}{c|}{\multirow{2}{*}{Methods}} & \multicolumn{2}{c|}{\multirow{1}{*}{PolitiFact}} & \multicolumn{2}{c}{GossipCop}  \\ 
            & Accuracy & Macro-F1 & Accuracy & Macro-F1  \\ \hline 
               
            \multicolumn{1}{l|}{{\bf PSS-TL}} &\textbf{93.23} & \textbf{92.47}&\textbf{97.71} &\textbf{97.62} \\
            \multicolumn{1}{l|}{{\quad w/o Content Teacher}} & 90.32&89.29 &97.44 &97.32  \\ 
            \multicolumn{1}{l|}{{\quad w/o Propagation Teacher}} &91.94 &90.91 & 97.25&97.13 \\ 
             \multicolumn{1}{l|}{{\quad w/o $L_\text{tar}$}} & 91.94&90.91 &97.25 &97.18  \\ 
             \multicolumn{1}{l|}{{\quad w/o $L_\text{sup}$}} &88.71 &86.27 & 97.16& 97.03 \\             
            \multicolumn{1}{l|}{{\quad w/o LGPI}} & 90.32&88.89 &97.20 &97.19  \\ \hline 
 
        \end{tabular}
\end{table}

The results are shown in Table \ref{tab:ablation}.
Our full PSS-TL achieves better performance on both datasets.
It can be observed that: 1) 
On PolitiFact and GossipCop, removing the propagation teacher or content teacher leads to a more pronounced decrease in performance, which indicates that the two teachers are capable of effectively capturing semantics and propagation characteristics within noisy data respectively.
 2) When removing teacher supervision loss, our method achieves worse performance on both evaluation metrics, demonstrating the usefulness of teachers' supervision to the student. 3) Meanwhile, the performance of w/o L-tar is significantly lower than PSS-TL, validating our hypothesis that targeted guidance can provide additional knowledge supplementation for the student. 4) As the student model relies solely on local propagation information for detection, w/o LGPI is inferior to PSS-TL, confirming the effectiveness of the LGPI module in our method.

\subsection{Generalization Evaluation}

\begin{table}[t]
    \centering
        \caption{Performance comparison of PSS-TL and DECOR in cross-domain detection. \textit{PolitiFact $\longrightarrow$ \text{COAID}} refers to training on the source domain (i.e., PolitiFact) and testing on the target domain (i.e., COAID).}
    \label{tab:cross}
  
        \begin{tabular}{l|cc|cc|cc}
            \hline 
            \multicolumn{1}{c|}{\multirow{2}{*}{Methods}} & \multicolumn{2}{c|}{\multirow{1}{*}{PolitiFact $\longrightarrow$ COAID}} & \multicolumn{2}{c|}{GossipCop $\longrightarrow$ COAID} & \multicolumn{2}{c}{COAID $\longrightarrow$ COAID} \\ 
            & Accuracy& Macro-F1 & Accuracy& Macro-F1 & Accuracy& Macro-F1 \\ \hline 
            \multicolumn{1}{l|}{{DECOR}} &89.47 & 93.94&84.21 &90.91&97.37&93.70  \\
            \multicolumn{1}{l|}{{UPSR}} & 84.21&91.43 &73.68&68.42 &94.74&97.06 \\            
            \multicolumn{1}{l|}{\textbf{PSS-TL}} & \textbf{92.11}&\textbf{95.65} &\textbf{89.47} &\textbf{94.12} &\textbf{100.00}&\textbf{100.00} \\ 
     \hline     
        \end{tabular}

\end{table}

\begin{figure}[t]
\centering

  \subfigure[PolitiFact]{\includegraphics[width=0.4\linewidth]{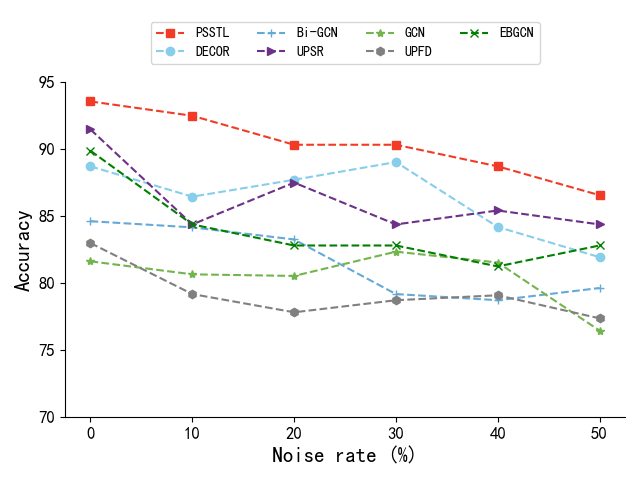}}
  \subfigure[GossipCop]{\includegraphics[width=0.4\linewidth]{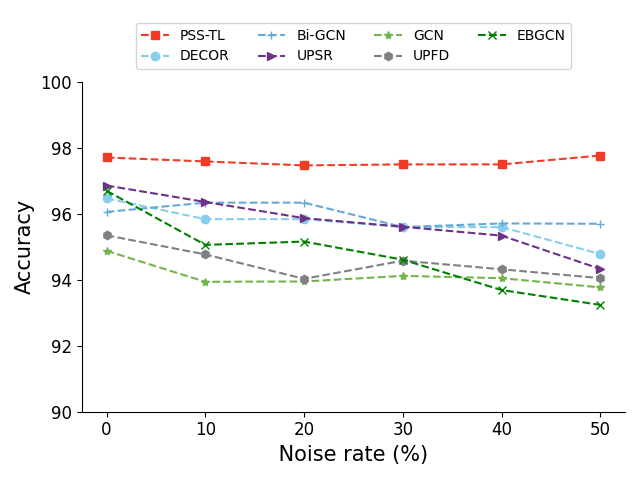}}
\caption{Robust detection results (\%) against different ratios of semantic noises.}
  \label{fig:noise_semantic}
\end{figure}

\begin{figure}[t]
\centering
  \subfigure[PolitiFact]{\includegraphics[width=0.4\linewidth]{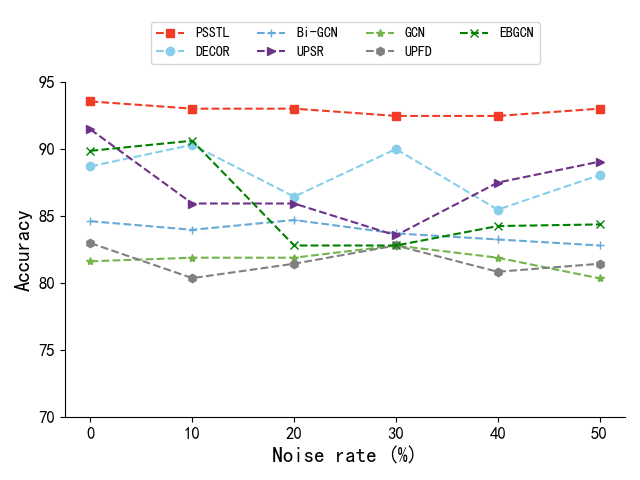}}
  \subfigure[GossipCop]{\includegraphics[width=0.4\linewidth]{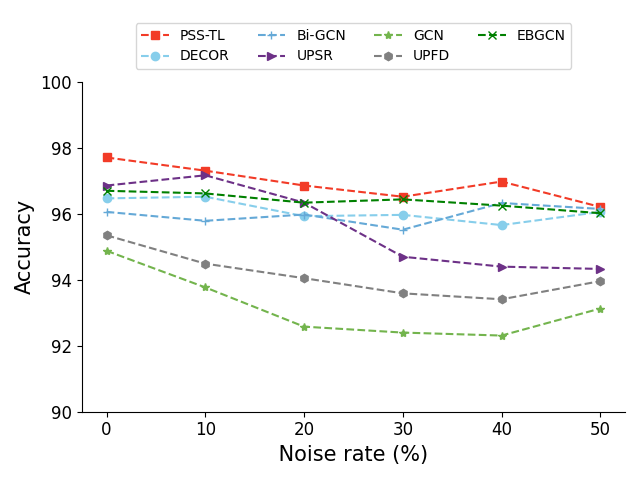}}
\caption{Robust detection results (\%) against different ratios of structural noises.}
  \label{fig:noise_pro}
\end{figure}

\begin{figure}[!t]
\centering
 \subfigure[PolitiFact]{\includegraphics[width=0.4\linewidth]{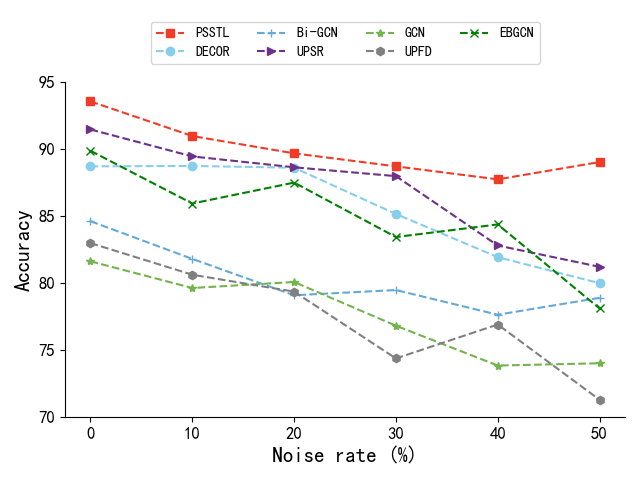}}
  \subfigure[GossipCop]{\includegraphics[width=0.4\linewidth]{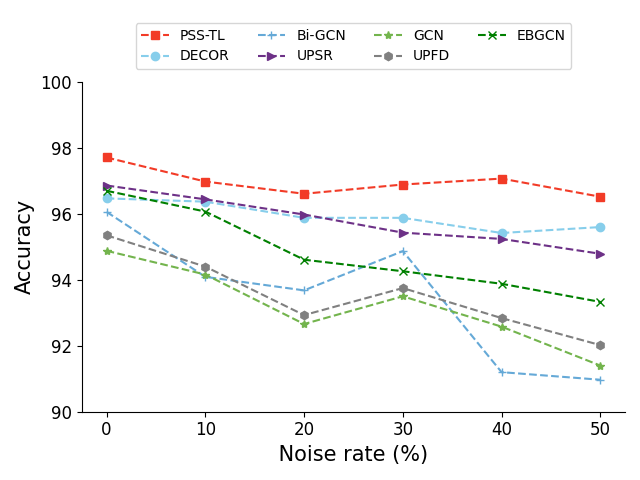}}
\caption{Robust detection results (\%) against different ratios of mixed noises (i.e., semantic and structural noises).}
  \label{fig:mix_noise}
\end{figure}

\begin{figure}[!t]
\centering
 \subfigure[PolitiFact]{\includegraphics[width=0.4\linewidth]{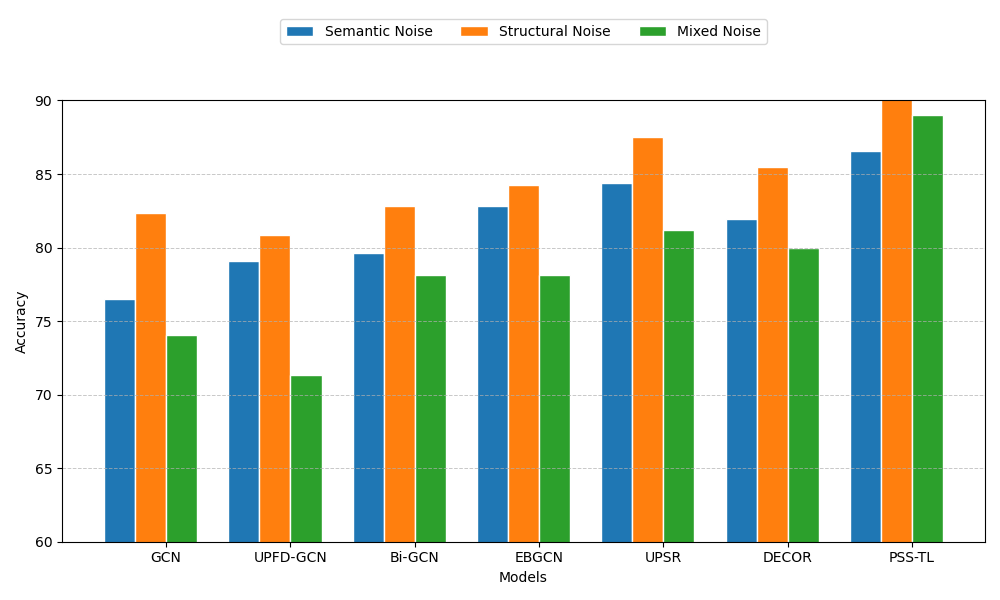}}
  \subfigure[GossipCop]{\includegraphics[width=0.4\linewidth]{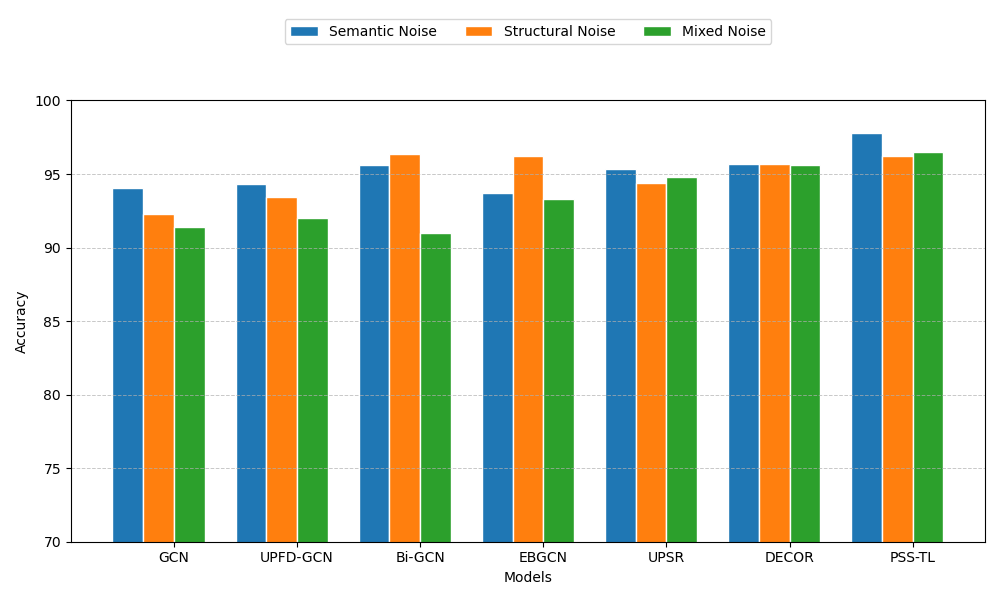}}
\caption{Robust detection results (\%) against different models with different noise (The ratio of noise is 0.5).}
  \label{fig:mutual_inf}
\end{figure}

We conduct cross-domain fake news detection to evaluate the generalization of our method.
The results are shown in Table \ref{tab:cross} where GCN is adopted as the GNN backbone for PSS-TL and DECOR.  

The PSS-TL method exhibits remarkable performance, outperforming other techniques. Specifically, compared to DECOR, PSS-TL achieves a 6.25\% enhancement in accuracy when transferring from GossipCop to COAID, and a 2.95\% improvement when transferring from PolitiFact to COAID. The results indicate that PSS-TL has a kind of generalization ability, which can be attributed to its effective transfer learning mechanisms. 
Moreover, our method simultaneously models the structural information of user interactions and the semantic information of content, thereby obtaining more general and more discriminative detection features for cross-domain detection.

\subsection{Robustness Evaluation}

To further validate the effectiveness and robustness of our method under three noisy scenarios, single semantic noises, single structural noises and mixed noises. we introduce random masking to the node features and edges in the news propagation graph and conduct tests on datasets with different ratios of noise.

 Figs.~\ref{fig:noise_semantic}, \ref{fig:noise_pro} and \ref{fig:mix_noise} show the results of different methods under semantic, structural and mixed noises, respectively.
Our method consistently outperforms other methods and exhibits more stable performance under different noisy scenarios and noise rates. 
In the semantic noise scenario, PSS-TL performs relatively better on PolitiFact and even maintains its performance on GossipCop, whereas some propagation-based methods such as Bi-GCN suffer significantly on PolitiFact. 
In the structural noise scenario, some methods such as UPSR may benefit from the structural noise but still perform less than PSS-TL. 
As shown in Fig.\ref{fig:mix_noise} and Fig.\ref{fig:mutual_inf}, for mixed noise scenarios, some methods for addressing propagation noise have faced challenges, for instance, when the noise ratio is 0.5 on PolitiFact, EBGCN, UPSR, and DECOR have been subjected to more severe challenges than the other two types of noise scenarios. This discrepancy might be attributed to the mutual interference between semantic noises and structural noises. Nevertheless, our method still suffers less in such a scenario, which validates the effectiveness and robustness of our method under
 various scenarios. 

\subsection{Parameter Analysis}
\begin{figure}[t]
    \centering  
      \subfigure[PolitiFact]{\includegraphics[width=0.49\linewidth]{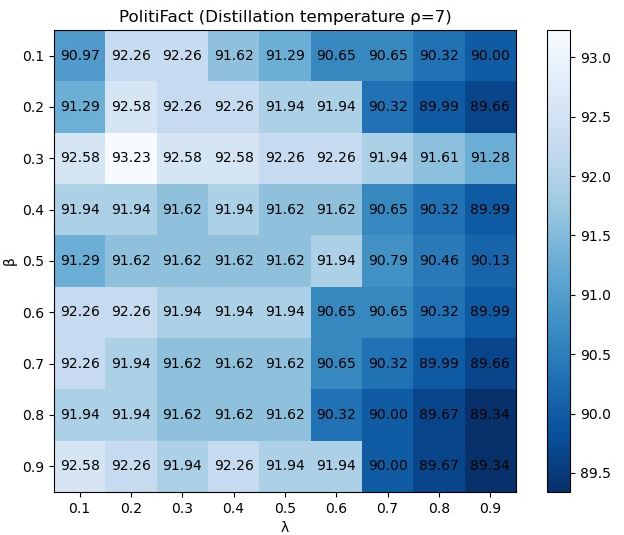}}
      \subfigure[GossipCop]{\includegraphics[width=0.49\linewidth]{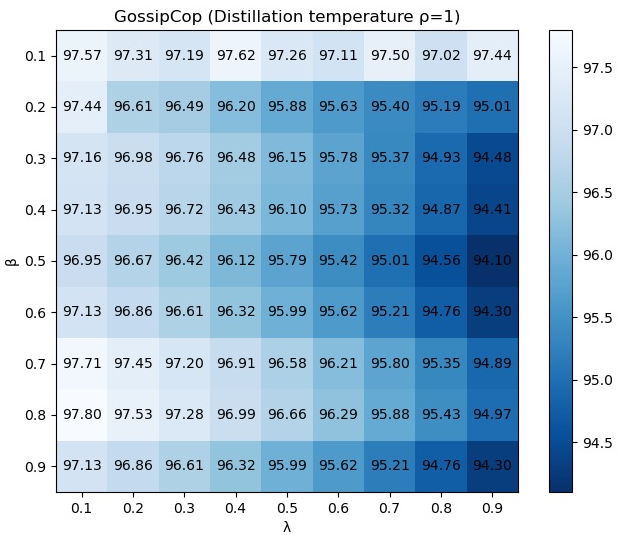}}
    \caption{Parameter analysis of $\lambda$ and $\beta$.}
    \label{fig:para_analysis}
\end{figure}
\begin{figure}[t]
    \centering  
      \subfigure[PolitiFact]{\includegraphics[width=0.4\linewidth]{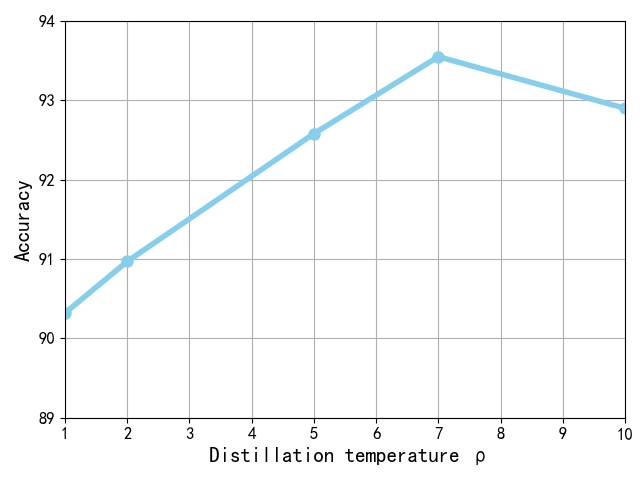}}
      \subfigure[GossipCop]{\includegraphics[width=0.4\linewidth]{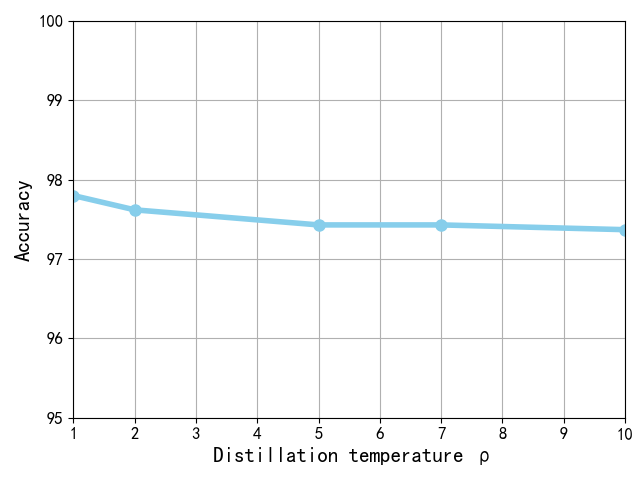}}
    \caption{Parameter analysis of $\rho$.}
    \label{fig:para2_analysis}
\end{figure}

We explore the performance of PSS-TL against three vital parameters, $i.e.$, the two parameters to balance the influence of different
teachers in teacher supervision ($\lambda$) and targeted guidance ($\beta$), and the temperature parameter in teacher supervision ($\rho$). Fig.~\ref{fig:para_analysis} and Fig.~\ref{fig:para2_analysis} show the results of PSS-TL with different values of $\lambda$ and $\beta$.

\textbf{Effect of $\lambda$}. The larger 
$\lambda$ is, the more the student model tends to be supervised by the propagation teacher. The best settings are 0.1 and 0.2 on GossipCop and PolitiFact, respectively. This is due to the greater impact of noise with more nodes in the graph, necessitating more guidance from the propagation teacher.

\textbf{Effect of $\beta$}. $\beta$ represents the extent to which the propagation teacher supplements the student model’s knowledge in targeted guidance. The optimal setting is 0.8 and 0.3 on GossipCop and PolitiFact, respectively. The differences between
$\beta$ and $\lambda$ highlight the necessity of weighing guidance for different teachers in targeted instruction.

\textbf{Effect of $\rho$}.   The temperature parameter $\rho$ adjusts the smoothness of the output generated by teacher models. The larger the $\rho$, the student model will receive more knowledge from the teacher model. we fix the best settings for $\lambda$ and $\beta$. The optimal setting is 1 and 7 on GossipCop and PolitiFact, respectively. It suggests that on the PolitiFact dataset, student models need to acquire more knowledge because the interaction between content noise and propagation noise has a greater impact on this dataset.

\section{Conclusion}
This paper investigates the issue of mutual interference between feature noise and propagation structural noise in fake news detection. We propose a multi-channel teacher-student knowledge transfer method, which enables the student model to learn content knowledge and propagation knowledge of news separately based on the guidance of the teacher model, without being interfered with by the noise of both. Experiments on two real-world datasets demonstrate the effectiveness of the proposed method.

\section*{Acknowledgements}
This work was supported by the National Key Research and Development Program of China (No. 2022YFC3302102), and the Postdoctoral Fellowship Program of CPSF (No. GZC20232969).
\bibliographystyle{splncs04}

\end{document}